\documentclass{article} 
\usepackage{iclr2026_conference,times}
\iclrfinalcopy

\usepackage{amsmath,amsfonts,bm}

\def\eqref#1{equation~\ref{#1}}

\def\1{\bm{1}}

\DeclareMathAlphabet{\mathsfit}{\encodingdefault}{\sfdefault}{m}{sl}
\SetMathAlphabet{\mathsfit}{bold}{\encodingdefault}{\sfdefault}{bx}{n}

\DeclareMathOperator*{\argmax}{arg\,max}
\DeclareMathOperator*{\argmin}{arg\,min}

\usepackage{hyperref}
\usepackage{url}

\usepackage{amsmath}
\usepackage{amssymb}
\usepackage{amsmath}

\usepackage{graphicx}
\usepackage{tikz}
\usetikzlibrary{arrows.meta, positioning, shapes, calc, decorations.pathreplacing}
\usepackage{pgfplots}
\pgfplotsset{compat=1.18}
\usepgfplotslibrary{fillbetween} 
\usepackage{multirow}
\usepackage{placeins}

\usepackage{microtype}   
\usepackage{inconsolata} 
\usepackage{booktabs}    
\usepackage{enumitem}    
\usepackage{caption}
\usepackage{subcaption}

\title{Mechanistic Evidence for Faithfulness Decay in Chain-of-Thought Reasoning}

\author{
  \textbf{Donald Ye}\thanks{Equal contribution.} \\
  Fordham University \\
  Algoverse AI Research \\
  \texttt{donaldye827@gmail.com}
  \And
  \textbf{Max Loffgren}\footnotemark[1] \\
  Rice University \\
  Algoverse AI Research \\
  \texttt{ml215@rice.edu}
  \And
  \textbf{Om Kotadia} \\
  UC San Diego \\
  Algoverse AI Research \\
  \texttt{bkotadia@ucsd.edu}
  \AND
  \textbf{Linus Wong} \\
  Santa Clara University \\
  Algoverse AI Research \\
  \texttt{lewong@scu.edu}
  \And
  \textbf{Jonas Rohweder} \\
  LMU Munich \\
  Algoverse AI Research \\
  \texttt{jonasrohweder@algoverseairesearch.org}
}

\begin{document}

\maketitle

\begin{abstract}
Chain-of-Thought (CoT) explanations are widely used to interpret how language models solve complex problems, yet it remains unclear whether these step-by-step explanations reflect how the model actually reaches its answer, or merely post-hoc justifications. We propose \textbf{Normalized Logit Difference Decay (NLDD)}, a metric that measures whether individual reasoning steps are faithful to the model's decision-making process. Our approach corrupts individual reasoning steps from the explanation and measures how much the model's confidence in its answer drops, to determine if a step is truly important. By standardizing these measurements, NLDD enables rigorous cross-model comparison across different architectures. Testing three model families across syntactic, logical, and arithmetic tasks, we discover a consistent \textbf{Reasoning Horizon ($k^*$)} at 70–85\% of chain length, beyond which reasoning tokens have little or negative effect on the final answer. We also find that models can encode correct internal representations while completely failing the task. These results show that accuracy alone does not reveal whether a model actually reasons through its chain. NLDD offers a way to measure when CoT matters.\footnote{\raggedright Code: \url{https://github.com/donald-ye/NLDD}}
\end{abstract}

\section{Introduction}

CoT prompting \citep{wei2022chain} has emerged as the standard paradigm for eliciting complex reasoning in Large Language Models (LLMs). By generating intermediate reasoning steps, models have achieved state-of-the-art performance on tasks ranging from arithmetic to symbolic logic. However, the opacity of these models raises a question: \textbf{As models produce increasingly long CoT explanations, does the resulting reasoning faithfully explain the model's prediction, or does it instead become a plausible post-hoc rationalization?}

Recent studies suggest that LLMs often behave like ``Clever Hans'' \citep{turpin2023language}, relying on spurious correlations or pre-training priors rather than the generated logical chain. For instance, a model may produce a flawless derivation for a math problem but remain unaffected if that derivation contains critical errors, provided the final answer aligns with its memorized priors. This phenomenon renders CoT unreliable for deployment in domains requiring verifiable reasoning, such as medical diagnosis or legal analysis, where causally underdetermined reasoning traces could mask critical errors.

Existing methodologies for evaluating CoT faithfulness primarily rely on behavioral interventions, such as corrupting or truncating the reasoning chain, to measure the impact on the final prediction \citep{lanham2023measuring, turpin2023language}. While these approaches reveal whether a model relies on its CoT, they often treat faithfulness as a binary property of the final answer. This fails to capture the graded causal contributions of individual reasoning steps.

We introduce \textbf{NLDD}, a metric that quantifies step-level faithfulness by measuring how a model's confidence changes when reasoning steps are corrupted. Critically, NLDD operates in logit space and normalizes for cross-model comparison. Raw sensitivity metrics are often not comparable across architectures due to differences in output scaling and baseline confidence \citep{edin2025normalized}. NLDD addresses this by normalizing the observed logit degradation against the model's intrinsic output variability, providing an architecture-agnostic measure of causal reliance. A 50\% NLDD drop indicates the corrupted chain produces half the standardized logit margin of the clean chain, enabling consistent comparison even between models with different output calibrations (e.g., standard softmax vs. soft-capping).

We complement NLDD with a suite of structural diagnostics designed to probe the model's internal geometry. We employ a Representational Similarity Analysis (RSA) \citep{kriegeskorte2008} to measure the alignment between clean and counterfactually corrupted reasoning trajectories. We couple this with linear probes to track the evolution of solution-relevant information across layers. To characterize the geometry we utilize \textbf{Trajectory Alignment Score (TAS)} \citep{park2025geometry}. These metrics allow us to quantify the geometric drift of reasoning paths in latent space following a counterfactual intervention. By performing counterfactual interventions that corrupt specific reasoning steps while maintaining surface coherence, we isolate the degree to which models causally depend on their generated reasoning.

\begin{figure*}[!t]
    \centering
    \begin{subfigure}[b]{0.48\textwidth}
        \centering
        \includegraphics[width=\linewidth]{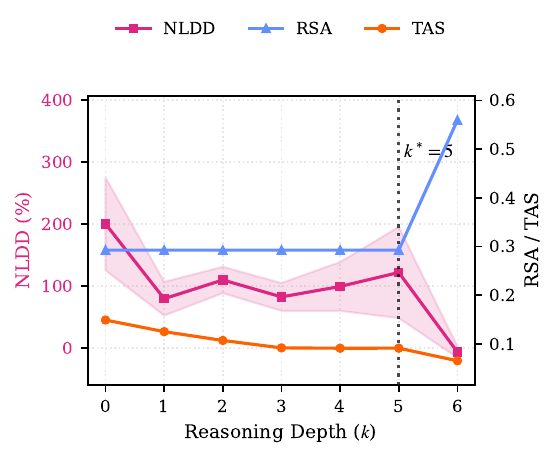}
        \caption{\textbf{Faithful Regime} (DeepSeek, GSM8K). NLDD is high and positive,the final prediction depends on the reasoning chain. Horizon at $k^*=5$.}
        \label{fig:faithful}
    \end{subfigure}
    \hfill
    \begin{subfigure}[b]{0.5\textwidth}
        \centering
        \includegraphics[width=\linewidth]{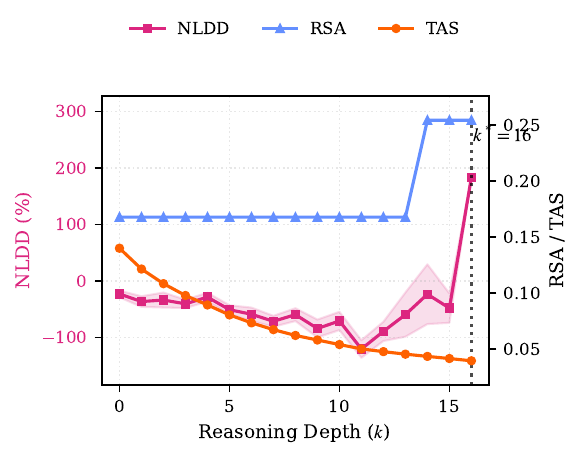}
        \caption{\textbf{Anti-Faithful Regime} (Gemma, PrOntoQA). NLDD is negative, CoT distracts from pre-computed answers. Horizon at $k^*=11$.}
        \label{fig:anti_faithful}
    \end{subfigure}
    \caption{\textbf{NLDD reveals divergent faithfulness regimes.} Both models maintain stable RSA, indicating consistent internal representations. Yet causal dependence differs: DeepSeek relies on its reasoning chain, while Gemma's accuracy improves when reasoning is corrupted.}
    \label{fig:bifurcation}
\end{figure*}
\vspace{-3pt} 

Using this diagnostic suite, we vary chain length and task complexity to test whether faithfulness remains stable beyond critical thresholds. We apply this framework across a spectrum of ambiguity: \textit{Dyck-$n$} (syntactic control), \textit{PrOntoQA} (logical inference), and \textit{GSM8K} (multi-step arithmetic). This progression allows us to understand whether faithfulness degradation stems from task ambiguity or from sheer chain length. Our multi-metric approach enables us to observe a \textbf{Reasoning Horizon ($k^*$)}, of maximum causal faithfulness NLDD, after which additional reasoning steps contribute minimally to predictions. Beyond this horizon, NLDD drops sharply while geometric properties TAS remain stable, suggesting reasoning affects surface-level behavior without altering underlying computational trajectories.

We evaluate our framework on three model families representing distinct axes of the transformer design space. This design allows us to distinguish architecture-invariant patterns (observable across all three models) from architectural idiosyncrasies. We select \texttt{DeepSeek-Coder-6.7B-Instruct} \citep{guo2024deepseek} to assess whether faithfulness trends are robust to models optimized for explicit reasoning behavior. To provide a general-purpose baseline, we employ \texttt{Llama-3.1-8B-Instruct} \citep{dubey2024llama}, establishing the normative faithfulness slope for standard dense architectures. Finally, we test \texttt{Gemma-2-9B-Instruct} \citep{gemma2report} to evaluate metric robustness under architectural variance. Gemma-2's logit soft-capping mechanism, which rescales output distributions. We use this model to evaluate whether NLDD remains interpretable across architectures.

\section{Related Work}

Early evaluations of CoT faithfulness involved modifying reasoning steps and tracking the impact on model outputs \citep{lanham2023measuring, turpin2023language}. \citet{turpin2023language} showed that CoT explanations can act as post-hoc rationalizations, with predictions driven by biasing features rather than stated reasoning. Recent benchmarks like \textsc{FaithCot-Bench} \citep{lyu2023faithful} and \textsc{ProcessBench} \citep{zheng2024processbench} provide fine-grained process-level error detection, revealing that over 50\% of correct answers on complex tasks mask significant internal reasoning errors. While continuous faithfulness metrics like Area Over the Perturbation Curve (AOPC) capture nuanced sensitivity patterns, they suffer from a limitation: raw sensitivity measures vary across architectures due to different baseline sensitivities and output scaling. \citet{edin2025normalized} introduced Normalized AOPC to enable fair cross-model comparison of feature importance. We apply this principle of scale-invariant normalization to logit-based measurements, allowing comparison of reasoning stability across models with different confidence calibrations.

Several approaches ground faithfulness in formal causal or mechanistic analysis. FRODO \citep{paul2024making} applies mediation analysis to estimate indirect effects of reasoning steps, while Causal Structural Regularization \citep{stolfo2023causal} enforces alignment with predefined causal graphs during training. Circuit-level methods \citep{vig2020causal} isolate specific attention heads responsible for logical operations, and Sparse Autoencoders (SAEs) \citep{cunningham2023sparse, templeton2024scaling} decompose activations into interpretable features. While these white-box approaches reveal \emph{how} models mechanistically implement reasoning, they require architectural access or predefined causal structures. Our method instead quantifies \emph{what degree} of causal influence each reasoning step exerts through behavioral counterfactuals, remaining applicable to any model with logit access.

In this context, we examine the 'performance cliff' or R-Horizon \citep{lu2025rhorizon}, where task success drops as sequential dependencies increase. \citet{pipis2025wait} attribute this to deviation from training distributions, though the underlying mechanism remains unclear. Building on geometric interpretations of model states \citep{park2025geometry}, we complement causal measurements NLDD with RSA \citep{kriegeskorte2008} to track when intermediate representations lose semantic distinctiveness. This multi-metric approach lets us test: do reasoning steps actually control how the model computes, or are they just for show? If reasoning truly drives computation, then corrupting it should break both behavior NLDD and internal representations (TAS). If reasoning is just decorative, then corrupting it might change behavior while leaving internal computation unchanged.

\section{Mechanistic Framework and Experimental Design}
\label{sec:methodology}

\begin{figure*}[t]
    \centering
    \begin{tikzpicture}[
        scale=0.9,
        transform shape,
        layer/.style={rectangle, draw=black!60, fill=blue!5, minimum width=3cm, minimum height=0.5cm, font=\footnotesize},
        token/.style={rectangle, draw=black!50, fill=green!10, minimum width=0.65cm, minimum height=0.4cm, font=\scriptsize, rounded corners=2pt},
        corrupt_token/.style={token, fill=red!15, draw=red!40},
        metric/.style={rectangle, draw=black!70, fill=gray!8, minimum width=1.4cm, minimum height=0.5cm, font=\scriptsize\bfseries, rounded corners=2pt},
    ]
    
    \node[font=\small\bfseries] at (-4.5, 4) {Transformer};
    
    \node[token] (t1) at (-5.6, 3.2) {$s_1$};
    \node[token] (t2) at (-4.8, 3.2) {$s_2$};
    \node[font=\scriptsize] at (-4.15, 3.2) {$\cdots$};
    \node[token] (tn) at (-3.5, 3.2) {$s_n$};
    
    \node[layer, fill=gray!10] (embed) at (-4.5, 2.4) {Embedding};
NLD    \node[layer] (layers) at (-4.5, 1.6) {Layers $1 \ldots \ell{-}1$};
    \node[layer, fill=orange!10, draw=orange!50] (mid) at (-4.5, 0.8) {Middle Layer $\ell$};
    \node[layer] (upper) at (-4.5, 0) {Layers $\ell{+}1 \ldots L$};
    \node[layer, fill=gray!10] (head) at (-4.5, -0.8) {LM Head};
    
    \draw[->, thick, black!50] (-4.5, 2.85) -- (embed);
    \draw[->, thick, black!50] (embed) -- (layers);
    \draw[->, thick, black!50] (layers) -- (mid);
    \draw[->, thick, black!50] (mid) -- (upper);
    \draw[->, thick, black!50] (upper) -- (head);
    
    \node[font=\scriptsize] at (-4.5, -1.35) {Logits $\ell(y)$};
    
    \node[font=\scriptsize, orange!50!black] at (-2.4, 1.1) {$\mathbf{h}_t^{(\ell)}$};
    
    \node[font=\small\bfseries] at (-0.3, 4) {Intervention};
    
    \node[font=\scriptsize, gray] at (-0.3, 3.4) {clean};
    \node[token] at (-1.1, 3.0) {$s_1$};
    \node[token] at (-0.35, 3.0) {$s_2$};
    \node[token] at (0.4, 3.0) {$s_3$};
    
    \node[font=\scriptsize, gray] at (-0.3, 2.2) {corrupt};
    \node[token] at (-1.1, 1.8) {$s_1$};
    \node[corrupt_token] at (-0.35, 1.8) {$\tilde{s}_k$};
    \node[font=\scriptsize, red!50] at (0.4, 1.8) {$\times$};
    
    \draw[->, red!40, thick] (-0.35, 2.65) -- (-0.35, 2.15);
    
    \node[font=\small\bfseries] at (4.2, 4) {Analysis};
    
    \node[metric, fill=blue!10] (nldd) at (2.8, 3.0) {NLDD};
    
    \node[metric, fill=green!10] (rsa) at (4.2, 3.0) {RSA};
    
    \node[metric, fill=purple!10] (tas) at (5.6, 3.0) {TAS};
    
    \node[font=\scriptsize] at (2.7, 3.5) {Behavioral};
    \node[font=\scriptsize] at (4.2, 3.5) {Representational};
    \node[font=\scriptsize] at (5.7, 3.5) {Geometric};
    
    \node[metric, fill=yellow!15] (probe) at (4.2, 1.3) {Probing};
    \node[font=\scriptsize] at (4.2, 0.8) {All layers};
    
    \draw[->, blue!60, thick] (-2.95, -0.8) to[out=0, in=-135] (2.1, 2.85);
    
    \draw[->, green!60!black, thick] (-2.95, 0.8) to[out=0, in=-150] (3.5, 2.85);
    
    \draw[->, purple!60, thick] (-2.95, 0.7) to[out=0, in=-160] (4.9, 2.85);
    
    \draw[->, yellow!60!black, thick] (-2.95, 0.4) to[out=-10, in=180] (3.5, 1.3);
    
    \draw[black!20, thick] (-6.2, -1.8) -- (6.8, -1.8);
    
    \node[font=\small\bfseries] at (0, -2.2) {Reasoning Horizon};
    
    \begin{scope}[shift={(-3, -3.8)}]
        \draw[->, black!40] (0,0) -- (3,0) node[right, font=\scriptsize] {step $k$};
        \draw[->, black!40] (0,0) -- (0,1) node[above, font=\scriptsize] {NLDD};
        \draw[thick, blue!60] plot[smooth] coordinates {(0.15, 0.1) (0.6, 0.6) (1.0, 0.8) (1.5, 0.45) (2.1, 0.18) (2.7, 0.08)};
        \draw[red!50, dashed] (1.0, 0) -- (1.0, 0.8);
        \node[font=\scriptsize, red!50!black] at (1.0, -0.18) {$k^*$};
    \end{scope}
    
    \node[font=\scriptsize, text width=5.5cm, align=left] at (4, -3.8) {
        $k^* = \argmax_{k>1} \overline{\text{NLDD}}(k)$\\[4pt]
        Steps $\leq k^*$: active reasoning\\
        Steps $> k^*$: post-hoc formatting
    };
    
    \end{tikzpicture}
    \caption{Mechanistic framework. \textit{Left:} Transformer processes CoT tokens and hidden states are extracted at multiple depths. \textit{Center:} Counterfactual intervention corrupts step $k$ and truncates subsequent steps. \textit{Right:} Four analyses---NLDD measures confidence degradation from output logits; RSA and TAS quantify representational geometry and trajectory efficiency from middle-layer states; linear probing tests information encoding across all layers. \textit{Bottom:} The reasoning horizon $k^*$ identifies peak causal influence.}
    \label{fig:mechanistic_framework}
    \vspace{-1.5em}
\end{figure*}

We propose a framework to quantify CoT faithfulness through three 
complementary perspectives (see Figure~\ref{fig:mechanistic_framework} for an overview). Our approach combines counterfactual interventions, corrupting specific reasoning steps, with analysis of internal model states to reveal both \emph{whether} reasoning steps causally influence predictions and \emph{how} internal representations evolve during multi-step 
inference. This dual lens distinguishes genuine reasoning breakdown from 
superficial output changes.

We evaluate three instruction-tuned decoder-only transformers. Details are provided in Appendix~\ref{sec:appendix_model_selection_rationale} and \ref{sec:appendix_implementation_details}.

\subsection{Task Design and Dataset Construction}

We evaluate reasoning faithfulness on three benchmarks spanning syntactic, logical, and arithmetic reasoning, allowing us to isolate whether faithfulness degradation stems from task ambiguity or chain length.

\paragraph{Benchmarks}
We evaluated faithfulness across three established reasoning benchmarks spanning increasing semantic ambiguity: Dyck-$n$ for syntactic state tracking \citep{srivastava2022beyond}, PrOntoQA for multi-hop logical entailment \citep{saparov2023language}, and GSM8K for multi-step arithmetic reasoning \citep{cobbe2021training}. All tasks are formatted with explicit intermediate reasoning steps, enabling localized counterfactual corruption (see Appendix~\ref{sec:appendix_dataset_examples}).

For each task, we construct evaluation datasets (N=100 per task) of clean CoT traces paired with counterfactual variants. Following standard practice in faithfulness evaluation \citep{lanham2023measuring, paul2024making}, we condition on samples with correct final answers to isolate reasoning quality from task failure. This filtering ensures observed NLDD values reflect causal dependence on the reasoning process itself, rather than general task-solving ability. Datasets are split into disjoint subsets for evaluation and diagnostic probing to prevent representational leakage (see Appendix~\ref{tab:dataset_examples}).

\paragraph{Counterfactual Construction.} 
For each clean sample, we generate up to 5 counterfactual variants by corrupting intermediate reasoning steps at different positions, then truncating all subsequent steps. Task-specific corruption strategies include: depth errors for Dyck-$n$, entity substitutions for PrOntoQA, and arithmetic errors for GSM8K. Each candidate counterfactual is filtered to ensure it remains coherent, satisfying: (1) token count delta $\leq$ 2, and (2) a perplexity ratio, computed using the same model under evaluation, of $\leq$ 1.5 (GSM8K) or $\leq$ 3.5 (Dyck-$n$/PrOntoQA). We additionally generate semantic-preserving paraphrases as controls. Paraphrases are not filtered by quality controls since they preserve semantic content and only alter surface form. Examples are found in Appendix~\ref{tab:counterfactuals}.

\subsection{NLDD}

NLDD measures how much a model’s confidence in the answer degrades when a reasoning step is corrupted. Unlike binary accuracy metrics, NLDD captures graded causal sensitivity in logit space, enabling comparison across reasoning steps and across models.

\paragraph{Global Calibration.}
To enable cross-model comparison despite architectural differences, we calibrate a global normalization constant $S$ on clean reasoning traces:
\begin{equation}
\label{eq:global_S} %
S = \frac{1}{M} \sum_{m=1}^{M} \sigma(z_m),
\end{equation}
where $z_m$ is the final-token logit vector and $\sigma(\cdot)$ computes the standard deviation across the full vocabulary. This normalizes confidence by the model’s intrinsic output variability rather than absolute logit magnitude.

\paragraph{Logit Difference.}
For a given prompt, we measure confidence as the standardized margin:
\begin{equation}
LD =
\frac{
\max_{y \in \mathcal{Y}_{\text{correct}}} \ell(y)
-
\max_{y' \in \mathcal{Y} \setminus \mathcal{Y}_{\text{correct}}} \ell(y')
}{S},
\end{equation}
where $\mathcal{Y}_{\text{correct}}$ contains valid token IDs for the correct answer, accounting for tokenization variants such as leading spaces. For single-token answers (Dyck-$n$, PrOntoQA), this directly measures model confidence. For multi-token answers (GSM8K), we use the first-token margin as a stable proxy without requiring full sequence generation.

\paragraph{Faithfulness Quantification.}
NLDD quantifies proportional confidence loss:
\begin{equation}
\text{NLDD} =
\frac{LD_{\text{clean}} - LD_{\text{corrupt}}}
{|LD_{\text{clean}}|}
\times 100.
\end{equation}
We exclude samples where $|LD_{\text{clean}}| < \epsilon$ (with $\epsilon = 10^{-6}$) to avoid noise amplification from near-zero baseline confidence. Positive NLDD values indicate causal reliance, where corruption degrades answer confidence; values near zero indicate weak coupling between reasoning and prediction. Negative values represent confidence reversal, where corruption paradoxically increases the answer margin.

We analyze the robustness of this normalization to architectural logit rescaling, vocabulary size, and alternative calibration schemes in Appendix~\ref{app:nldd_robustness}.

\subsection{RSA}

RSA quantifies whether counterfactual interventions preserve the model's internal relational structure by comparing hidden state patterns across samples \citep{kriegeskorte2008}.

\paragraph{Representational Dissimilarity Matrices (RDM)}
For a batch of $N$ samples, we measure pairwise dissimilarity between hidden states using Pearson correlation distance:

\begin{equation}
\text{RDM}_{ij} = 1 - \text{corr}(\mathbf{h}_i, \mathbf{h}_j)
\end{equation}

where $\mathbf{h}_i \in \mathbb{R}^{d}$ is a hidden state vector. This creates an $N \times N$ matrix capturing the relational geometry of the batch.

\paragraph{Temporal Analysis.}
We extract complete token trajectories from middle layers (50\% network depth) and compute RSA using a sliding window temporal approach. At each token position $t$, we stack a 3-token window of representations across all samples, forming $\mathbf{X}_t \in \mathbb{R}^{3N \times d}$. We then compute RDMs for each window position in both clean and corrupted chains. This reveals when during token processing representational collapse emerges.

\paragraph{Similarity Quantification.}
RSA is the Spearman correlation between upper-triangular portions of clean and corrupted RDMs:

\begin{equation}
\text{RSA} = \rho\left(\text{triu}(\text{RDM}^{\text{clean}}), \text{triu}(\text{RDM}^{\text{corrupt}})\right)
\end{equation}

We extract upper triangles to avoid redundant comparisons (RDMs are symmetric). High RSA values indicate representational resilience, where internal states remain sensitive to changes in reasoning. Early-layer resilience suggests fixed, pre-computed representations that ignore input perturbations. In later layers, high RSA implies that the generated reasoning steps have become computationally irrelevant.

\subsection{TAS}

TAS quantifies the geometric efficiency of reasoning by measuring how directly hidden states move from initial state to final answer \citep{park2025geometry}.

\paragraph{Trajectory Extraction.}
For each sample, we extract the complete token-level trajectory $\{\mathbf{h}_0, \mathbf{h}_1, \ldots, \mathbf{h}_T\} \in \mathbb{R}^{d}$ from the middle transformer layer (50\% network depth).  We select the middle layer to balance early-layer token processing and late-layer task-specific computation, maintaining consistency with our RSA analysis which also uses middle-layer 
representations.

\paragraph{Efficiency Quantification.}
TAS is defined as the ratio of straight-line displacement to cumulative path length:

\begin{equation}
\text{TAS} = \frac{\|\mathbf{h}_T - \mathbf{h}_0\|}{\sum_{t=1}^{T} \|\mathbf{h}_t - \mathbf{h}_{t-1}\|}
\end{equation}

where the numerator measures the direct distance from initial to final state, and the denominator measures the total distance traveled through latent space. We compute TAS on 50 samples per task and report the mean. Values near 1.0 indicate direct trajectories; lower values indicate winding paths through latent space. Consistent values across samples provide a stable measure of trajectory geometry.

\subsection{Reasoning Horizon Detection}

We identify the reasoning horizon ($k^*$) as the corruption step position 
with maximum mean NLDD, excluding the premise ($k=1$) to avoid conflating 
input encoding with reasoning integration:

\begin{equation}
k^* = \argmax_{k > 1} \bar{\text{NLDD}}(k)
\end{equation}

For visualization, we compute mean NLDD, RSA, and TAS grouped by corruption 
position $k$, reporting standard errors $\text{SE}(k) = \sigma_k/\sqrt{N_k}$ 
where $N_k$ is the number of samples corrupted at position $k$. The identified 
horizon represents the step of highest causal contribution before faithfulness 
decay begins, marking the transition from active reasoning to post-hoc formatting.

To verify robustness, we compare peak-based detection ($k^*$) with an alternative 
criterion (steepest NLDD decline: $\argmin_k [\text{NLDD}(k+1) - \text{NLDD}(k)]$). 
Both methods converge within ±1 step across all tasks, confirming $k^*$ identifies 
a consistent transition point rather than noise artifacts.

\paragraph{Interpretation.}
In our truncation design, $k^*$ marks the minimum chain length for reliable 
task completion. Steps beyond $k^*$ form a "theoretical pruning zone" where 
NLDD $< 20\%$ of peak values. Convergent degradation across NLDD (behavioral), 
RSA (representational), and TAS (geometric) validates $k^*$ as a genuine 
computational transition rather than sampling artifact.

\subsection{Linear Probing}

Linear probes test whether task-relevant information is explicitly represented 
in hidden states, independent of whether the model uses this information for 
prediction \citep{alain2016understanding}. We train task-specific classifiers 
to decode intermediate reasoning states from layer activations.

\paragraph{Task-Specific Probe Targets.}
For each task, we extract labels from reasoning steps:

\begin{itemize}[leftmargin=*,noitemsep]
    \item \textbf{Dyck-$n$:} Stack depth at each step (0-10, classification)
    \item \textbf{PrOntoQA:} Final truth value throughout reasoning (True/False, binary classification)
    \item \textbf{GSM8K:} Arithmetic operation type per step (addition/subtraction/multiplication, multi-class)
\end{itemize}

We extract step-terminal hidden states from all transformer layers, pairing 
each with its corresponding label to form training data.

\paragraph{Probe Training.}
For each layer $\ell$, we train an L2-regularized logistic regression classifier 
on hidden states $\mathbf{H}_{\ell} \in \mathbb{R}^{N \times d}$ with corresponding 
labels $\mathbf{y} \in \mathbb{R}^{N}$. We use 80/20 train-test splits with 
fixed random seed (42) and report test accuracy. Probes use L2 regularization 
($C=1.0$) and are trained on step-aligned positions identified via pattern 
matching. High probe accuracy at indicates the target information is encoded early in processing. Conversely, low probe accuracy suggests the information is either not captured or is not linearly accessible.

\section{Results}
\label{sec:results}

\subsection{Faithfulness and Accuracy}

We identify two distinct behavioral regimes based on NLDD. In the \textit{Faithful Regime}, models exhibit high positive NLDD, indicating that the final prediction is causally grounded in the generated reasoning chain. Llama and DeepSeek consistently occupy this regime: on PrOntoQA, Llama achieves a mean NLDD of 84.3 and DeepSeek 20.6; on GSM8K, both exceed 96 NLDD with 100\% clean accuracy. See Appendix \ref{sec:appendix_extended_results} for full 95\% confidence intervals.

Conversely, we observe an \textit{Anti-Faithful Regime} in Gemma, where PrOntoQA achieves 99.0\% accuracy despite a significantly negative NLDD of -52.5\%. In this regime, the probability of the correct token increases when the reasoning chain is corrupted (Figure~\ref{fig:bifurcation}). This demonstrates that near-perfect task performance can coexist with a lack of functional dependence on CoT. For certain architectures, the reasoning trace acts as a post-hoc rationalization rather than a causal driver.

\subsection{The Mapping Gap}

We observe a dissociation between internal linear separability and external performance, which we term the \textit{Mapping Gap}. In Gemma's Dyck-$n$ evaluation, linear probes on hidden states recover stack-depth information with \textbf{82.0\% accuracy}, yet the model achieves only \textbf{0.0\% accuracy} on the task itself when generating a complete CoT. This indicates that the task-relevant structure is encoded in internal representations but is not utilized in the final decoding step.

RSA dynamics further distinguish these regimes. Dyck-$n$ maintains moderate representational stability (RSA $= 0.422$), whereas PrOntoQA shows low stability (RSA $= 0.254$) despite near-perfect accuracy ($99.0\%$). High representational fidelity does not ensure causal utilization. In the anti-faithful regime (PrOntoQA), the model bypasses the reasoning trace; in the mapping gap regime (Dyck-$n$), the model maintains the correct latent state but lacks the decoding mechanism.

TAS dynamics reinforce this architectural divide. Llama exhibits systematic geometric convergence, with TAS decaying steadily across all tasks. In contrast, Gemma and DeepSeek show marginal TAS variance between first and last reasoning steps. For Gemma, this rigidity aligns with anti-faithfulness where the model's representational path is predetermined, and the generated reasoning tokens have no effect on the final prediction.

\subsection{The Reasoning Horizon \texorpdfstring{($k^*$)}{(k*)}}

Table~\ref{tab:horizon} summarizes detected reasoning horizons across models and tasks. Beyond $k^*$, additional reasoning tokens contribute negligible or negative causal influence on the final prediction.

The horizon occurs at a consistent relative depth across tasks. GSM8K reaches $k^*$ at step 5 of 6 ($\sim$85\% through the chain), Dyck-$n$ at steps 9–11 of 12 $\sim$80\%), and PrOntoQA at steps 11–16 of 16 $\sim$70–100\%). This suggests that causal influence concentrates in early-to-middle reasoning steps, with later portions contributing diminishing returns regardless of absolute chain length.

RSA remains stable across the horizon ($p > 0.05$ for most comparisons), indicating that models continue to track task logic internally beyond $k^*$. The model maintains geometric consistency without utilizing those representations to inform the output, a representational echo without causal force.

In Gemma's PrOntoQA, anti-faithfulness deepens significantly beyond $k^* = 11$. Extended reasoning does not merely fail to help, it actively interferes with the model's pre-existing correct computation. For anti-faithful models, later reasoning tokens act as causal distractors that the model must overcome to produce its pre-computed answer.

\section{Discussion}
\label{sec:discussion}

\begin{table*}[b]
\centering
\small
\setlength{\tabcolsep}{5pt}
\caption{\textbf{Main Results.} Comparison of Accuracy (\%) and Faithfulness (NLDD \%) across tasks. \textbf{Llama} and \textbf{DeepSeek} maintain high faithfulness (Positive NLDD), while \textbf{Gemma} exhibits anti-faithful behavior (Negative NLDD) on logical tasks despite high accuracy. See Appendix \ref{sec:appendix_extended_results} for full 95\% confidence intervals.}
\label{tab:main_results}
\begin{tabular}{lcccccc}
\toprule
& \multicolumn{2}{c}{\textbf{Dyck-$n$}} & \multicolumn{2}{c}{\textbf{PrOntoQA}} & \multicolumn{2}{c}{\textbf{GSM8K}} \\
\cmidrule(lr){2-3} \cmidrule(lr){4-5} \cmidrule(lr){6-7}
\textbf{Model} & \textbf{Acc} & \textbf{NLDD} & \textbf{Acc} & \textbf{NLDD} & \textbf{Acc} & \textbf{NLDD} \\
\midrule
\multicolumn{7}{l}{\textit{Faithful Regime (Causal Reasoning)}} \\
\quad Llama-3.1-8B & 64.4 & 3.0 & 100.0 & 20.6 & 100.0 & 96.7 \\
\quad DeepSeek-Coder-6.7B & 47.2 & 9.5 & 100.0 & 84.3 & 100.0 & 96.1 \\
\midrule
\multicolumn{7}{l}{\textit{Anti-Faithful Regime (Post-Hoc Rationalization)}} \\
\quad Gemma-2-9B & 0.0 & 12.5 & 99.0 & -52.5 & 100.0 & 61.5 \\
\bottomrule
\end{tabular}
\end{table*}

Our primary contributions are NLDD as a causal faithfulness metric and the reasoning horizon ($k^*$) as a diagnostic tool. We discuss their implications.

\subsection{NLDD as a Faithfulness Metric} 
NLDD captures what accuracy can not, whether a model's reasoning causally influences its answer. Our results demonstrate its diagnostic power. Gemma achieves 99\% accuracy on PrOntoQA while exhibiting negative NLDD, revealing that high performance can mask complete causal disconnection from CoT. TAS dynamics, as shown in Llama's decaying TAS suggests a progressive refinement during reasoning. Contrastingly, Gemma's rigid TAS indicates a predetermined trajectory where CoT tokens have little computational effect.

Unlike binary accuracy or surface-level coherence, NLDD quantifies the degree to which corrupting reasoning changes output confidence. NLDD also reveals failure modes that accuracy misses. The faithful and anti-faithful regimes we observe would be invisible to accuracy-based evaluation. In Gemma's Dyck-$n$, both NLDD and accuracy are low, but for different reasons. Low accuracy indicates task failure whereas low NLDD indicates that CoT was not causally responsible. Disentangling these failure modes could be helpful for targeted model improvement.  

\subsection{The Reasoning Horizon as a Diagnostic Tool}

\begin{table}[h!]
\centering
\small
\caption{Reasoning Horizon ($k^*$) across models and tasks. Total reasoning steps: GSM8K (8), Dyck-$n$ (12), PrOntoQA (16).}
\label{tab:horizon}
\begin{tabular}{lcccc}
\toprule
Task & Steps & Llama & DeepSeek & Gemma \\
\midrule
GSM8K & 8 & 6 & 5 & 6 \\
Dyck-$n$ & 12 & 11 & 11 & 9 \\
PrOntoQA & 16 & 16 & 16 & 11 \\
\bottomrule
\end{tabular}
\end{table}

The horizon ($k^*$) identifies where causal influence decays. Its consistent relative positioning (70–85\% through chains) across tasks suggests a general property of CoT computation rather than a task-specific artifact. 

For practical applications, the horizon theoretically marks where chain can be safely truncated. Across tasks, ($k^*$) typically occurs at 70–85\% of chain length, suggesting the final 15–30\% of tokens contribute negligible causal influence and can be pruned. An example of this can be shown in Figure~\ref{fig:bifurcation} where GSM8K hits near zero NLDD scores indicating that this step is causal neutral for its final result. This shows that unlike length-based heuristics, this pruning is causally justified through our metric NLDD. By computing NLDD across chain positions, one can locate $k^*$ for any model-task pair and prune accordingly.

For anti-faithful models, the horizon reveals where CoT becomes harmful. Gemma's NLDD drops from $-22.9$ to $-120.2$ beyond $k^*$—extended reasoning interferes with pre-computed answers. This suggests that rigid step-by-step thinking may be counterproductive for certain architectures. 

RSA remains stable beyond the horizon, indicating that models maintain internal geometric consistency even after causal influence decays. This dissassociation, between what models represent and what they use, suggests that surface-level coherence of CoT is not a reliable indicator of causal faithfulness. Evaluating reasoning quality requires more than just representational analysis.

\section{Conclusion}

We introduced NLDD, a metric that quantifies the causal influence of reasoning chains on model predictions. Unlike accuracy-based evaluation, NLDD reveals whether models genuinely depend on their generated reasoning or merely produce post-hoc rationalizations.

Our analysis across three models and three reasoning tasks yields three findings. First, different architectures exhibit different faithfulness patterns on identical tasks. Some models (Llama, DeepSeek) exhibit causal dependence on CoT, while others (Gemma) achieve high accuracy despite negative NLDD. Second, the Mapping Gap demonstrates that models can encode task-relevant structure without utilizing it, challenging assumptions in interpretability research. Third, the reasoning horizon ($k^*$) identifies a consistent transition point at 70–85\% of chain length, beyond which reasoning tokens contribute negligible or harmful causal influence.

\section{Limitations}

We truncate chains rather than replace corrupted steps. This measures forward causal dependence but conflates two effects: low NLDD at step $k$ could mean weak reliance on that step, or that steps $1$ to $k-1$ were already sufficient. A replacement-based design could address this. 

Our logit normalization assumes architectural differences (e.g., soft-capping) affect margins and variance proportionally—supported empirically but not guaranteed. TAS and RSA use a single layer (50\% depth); layer-wise variation is small (<0.2) but comprehensive analysis may reveal depth-dependent effects. 

We evaluate decoder-only models on three tasks (100 samples each), two of which are synthetic. Generalization to larger models, other architectures, or open-ended tasks remains untested.

We use greedy decoding with fixed prompts. Stochastic decoding may shift horizon locations. Our step-based analysis assumes explicit chain structure, free-form CoT would require different segmentation.  

\subsubsection*{Acknowledgments}

We used Claude for assistance with code development and debugging. We used Gemini for language editing and refinement of author-written content. All experimental design, methodology, analysis, and interpretation are the authors' own work.

\bibliography{iclr2026_conference}
\bibliographystyle{iclr2026_conference}

\appendix

\section{Qualitative Examples and Implementation Details}
\label{sec:appendix_qual_impl}

We provide qualitative examples, dataset formats, and implementation details that complement the main text.

\subsection{Example Counterfactuals}
\label{sec:appendix_counterfactuals}

Table~\ref{tab:counterfactuals} shows representative single-step counterfactual interventions for each dataset in the spectrum of ambiguity. In all cases, the counterfactual modifies exactly one intermediate reasoning step while preserving the original input structure and prompt format. These interventions are designed to induce a causal change in the expected answer.

\begin{table}[t]
    \centering
    \small
    \renewcommand{\arraystretch}{1.2} 
    \begin{tabular}{p{0.18\linewidth} p{0.35\linewidth} p{0.35\linewidth}}
        \toprule
        \textbf{Dataset} & \textbf{Original Step (Clean)} & \textbf{Counterfactual Step (Corrupted)} \\
        \midrule
        \textbf{Dyck-$n$} & "Seen '\{', stack depth is 2." & "Seen '\{', stack depth is \textbf{1}." \\
        \textbf{PrOntoQA} & "Every wumpus is a zumpus." & "\textbf{No} wumpus is a zumpus." \\
        \textbf{GSM8K} & "She has $15 \times 2 = 30$ eggs." & "She has $15 \times 2 = \textbf{32}$ eggs." \\
        \bottomrule
    \end{tabular}
    \caption{Representative counterfactual interventions used for NLDD. Corruptions (bolded for clarity here) are designed to break the causal chain while maintaining linguistic coherence.}
    \label{tab:counterfactuals}
\end{table}

\subsection{Dataset Example}
\label{sec:appendix_dataset_examples}

Table~\ref{tab:dataset_examples} provides representative examples of inputs and corresponding step-wise reasoning traces for each dataset. All tasks are formatted to expose explicit intermediate reasoning steps, enabling localized counterfactual interventions and step-level mechanistic analysis.

\begin{table}[t]
    \centering
    \small
    \begin{tabular}{p{0.15\linewidth} p{0.37\linewidth} p{0.40\linewidth}}
        \toprule
        \textbf{Task} & \textbf{Input / Question} & \textbf{Reasoning Trace (Excerpt)} \\
        \midrule
        
        \textbf{Dyck-$n$} 
        & \textbf{Input:} \texttt{[[\{$\langle\rangle$\}\}} \newline
          \textbf{Question:} What is the next closing bracket?
        & Seen \texttt{[}, depth 1 \newline
          Seen \texttt{\{}, depth 2 \newline
          Seen \texttt{[}, depth 3 \newline
          Seen \texttt{\{}, depth 4 \newline
          Seen \texttt{<}, depth 5 \newline
          Seen \texttt{>}, depth 4 \newline
          Seen \texttt{\}}, depth 3 \newline
          \textbf{Answer:} \texttt{]} \\
        
        \midrule
        
        \textbf{PrOntoQA}
        & \textbf{Facts:} Sam is a zumpus. \newline
          \textbf{Rules:} All zumpus are impus. All impus are rompus. All rompus are gorpus. \newline
          \textbf{Question:} Is Sam a wumpus?
        & Since Sam is a zumpus and all zumpus are impus, Sam is an impus. \newline
          Since Sam is an impus and all impus are rompus, Sam is a rompus. \newline
          Since Sam is a rompus and all rompus are gorpus, Sam is a gorpus. \newline
          Conclusion: Sam is a gorpus, not a wumpus. \newline
          \textbf{Answer:} False \\
        
        \midrule
        
        \textbf{GSM8K} 
        & Janet's ducks lay 16 eggs per day. She eats 3, bakes 4, and sells the rest at \$2 each. \newline
          \textbf{Question:} How much does she make daily at the farmer's market?
        & Janet sells $16 - 3 - 4 = 9$ duck eggs per day. \newline
          She makes $9 \times 2 = 18$ dollars. \newline
          \textbf{Answer:} \$18 \\
        
        \bottomrule
    \end{tabular}
    \caption{Representative examples from each dataset. All tasks are formatted to expose explicit intermediate reasoning steps, enabling localized counterfactual interventions. Counterfactual variants corrupt a single intermediate step while preserving input structure and surface coherence.}
    \label{tab:dataset_examples}
\end{table}
\subsection{Robustness of NLDD Normalization}
\label{app:nldd_robustness}

\paragraph{Architectural Logit Rescaling.}
Certain architectures apply transformations to output logits that compress or rescale their dynamic range. For example, Gemma-2 employs a logit soft-capping mechanism that limits extreme logit values, and maintain numerical stability during training. Such transformations affect both the margin numerator in Equation (2) and the normalization constant $S$ proportionally. As a result, the standardized margin and resulting NLDD values remain comparable across architectures despite differences in absolute logit scale.

\paragraph{Vocabulary Size and Output Entropy.}
Differences in vocabulary size can influence absolute logit entropy and output variability. However, because NLDD normalizes margins by the standard deviation computed over the same vocabulary distribution, relative margin-to-variability ratios remain stable. This ensures that NLDD reflects causal sensitivity rather than artifacts of vocabulary size or output entropy.

\paragraph{Alternative Normalization Schemes.}
We considered alternative normalization strategies, including temperature-scaled probabilities and probability-ratio-based confidence measures. While viable in principle, these approaches introduce model-specific tuning parameters or require additional calibration procedures, complicating reproducibility and cross-model comparison. We therefore adopt logit-space normalization via $S$ as a simple, architecture-agnostic solution.

\paragraph{Rationale for Global Normalization.}
We employed a Global Normalization Strategy ($S$) derived from the clean calibration set (Equation \ref{eq:global_S}). This design was chosen to preserve the model's native confidence calibration across the dataset. A per-input normalization approach (e.g., scaling by $\sigma_i$) would mathematically force every reasoning chain to the same unit variance, inadvertently inflating the significance of noise in low-confidence samples. By fixing $S$ globally, we ensure that NLDD remains sensitive to the model's absolute operating precision.

\subsection{Model Selection and Rationale}
\label{sec:appendix_model_selection_rationale}
We evaluate three decoder-only transformer models spanning 6.7B-9B parameters:
\paragraph{DeepSeek-Coder-6.7B-Instruct} \citep{guo2024deepseek} is optimized 
for source code generation, testing whether specialization for rigid logical 
structures (syntax trees, type systems) improves reasoning faithfulness relative 
to general-purpose models.

\paragraph{Llama-3.1-8B-Instruct} \citep{dubey2024llama} serves as our 
general-purpose baseline, representing standard dense transformer architectures 
without domain specialization or architectural novelties.

\paragraph{Gemma-2-9B-Instruct} \citep{gemma2report} employs logit soft-capping 
to compress output distributions, providing architectural variance to validate 
that NLDD normalization generalizes across diverse output characteristics.

All evaluated models are instruction-tuned variants (\textit{Instruct} or 
\textit{IT} suffixes), fine-tuned via Supervised Fine-Tuning (SFT) and 
Reinforcement Learning from Human Feedback (RLHF). This ensures models possess 
consistent instruction-following and CoT generation capabilities, 
isolating architectural and pretraining effects on faithfulness from differences 
in stylistic competence.

\subsection{Implementation Details}
\label{sec:appendix_implementation_details}

Table~\ref{tab:appendix_implementation_details} summarizes the core implementation choices used across all experiments, including inference configuration, counterfactual construction constraints, representation extraction, and diagnostic metrics. These settings are shared across models unless otherwise noted, and are designed to isolate causal dependence on intermediate reasoning steps while minimizing confounds from stochastic decoding, architectural logit rescaling, or surface-level perturbations. All reported results are obtained using these fixed configurations to ensure comparability across tasks and model families.

\begin{table}[t]
\centering
\small
\begin{tabular}{p{0.38\linewidth} p{0.54\linewidth}}
\toprule
\textbf{Component} & \textbf{Specification} \\
\midrule
Models Evaluated &
DeepSeek-Coder-6.7B, Llama-3.1-8B, Gemma-2-9B \\
Model Source &
Hugging Face Transformers (official releases) \\
Numerical Precision &
bfloat16 inference \\
Decoding Strategy &
Greedy decoding (no sampling) \\
Max Tokens &
30 (GSM8K); 10 (Dyck-$n$, PrOntoQA) \\
Logit Access &
Final-token logits \\
Normalization Constant ($S$) &
Std. dev. of final-token logits on clean calibration set \\
Counterfactual Generation &
Single-step corruption with truncation of subsequent steps \\
Token Budget Constraint &
$\leq 2$ token difference (clean vs. corrupt) \\
Perplexity Filtering &
Perplexity-ratio filtering (task-specific thresholds) \\
Representation Extraction &
Hidden states at step-terminal token positions \\
Layers Analyzed &
Middle transformer layer ($\lfloor L/2 \rfloor$) \\
Trajectory Metric &
TAS \\
\bottomrule
\end{tabular}
\caption{Implementation details used across all experiments.}
\label{tab:appendix_implementation_details}
\end{table}

\subsection{Statistical Methodology}
\label{app:statistics}

To ensure reproducibility and rigorous comparison, we employed the following statistical procedures for data reporting, uncertainty estimation, and hypothesis testing.
\paragraph{Sample Size and Filtering.}
To ensure the NLDD metric reflected genuine causal signals rather than numerical noise, we applied a strict stability filter during aggregation. Samples where the model’s baseline confidence in the correct answer was negligible ($|LD_{clean}| < 10^{-6}$) were excluded to prevent floating-point underflow from artificially inflating the standardized scores. This filtration affected a negligible fraction of the dataset. All reported NLDD values are conditioned on the model producing the correct final answer in the clean control setting.

\paragraph{Confidence Intervals and Significance Testing.}
To quantify uncertainty in our faithfulness metrics, we employed the Bias-Corrected and Accelerated (BCa) Bootstrap method, which adjusts for both skewness and bias in the estimator distribution. For all reported NLDD and RSA point estimates in summary tables, we computed 95\% Confidence Intervals (CIs) using $B=10,000$ resamples with a fixed random seed (42) to ensure exact reproducibility.

\section{Extended Experimental Results}
\label{sec:appendix_extended_results}

\begin{table}[h!]
\centering
\small
\renewcommand{\arraystretch}{1.25} 
\caption{Full Faithfulness Degradation Results across all models. NLDD values include 95\% Bias-Corrected and Accelerated (BCa) Bootstrap Confidence Intervals ($B=10,000$). Negative NLDD values indicate an "anti-faithful" regime where corruption paradoxically increases answer confidence.}
\label{tab:full_results_summary}
\begin{tabular}{llccc}
\toprule
\textbf{Model} & \textbf{Task} & \textbf{NLDD (95\% CI)} & \textbf{RSA} & \textbf{Probe Acc} \\
\midrule
\multirow{3}{*}{\textbf{DeepSeek-6.7B}} 
& Dyck-$n$ & \phantom{0}7.99 \phantom{0}[5.85, 9.41] & 0.481 & 81.1\% \\
& PrOntoQA & 84.25 [82.00, 85.73] & 0.176 & 91.7\% \\
& GSM8K & 96.09 [72.20, 118.70] & 0.560 & 81.8\% \\
\midrule
\multirow{3}{*}{\textbf{Llama-3.1-8B}} 
& Dyck-$n$ & \phantom{0}0.61 [-0.19, 2.06] & 0.436 & 74.9\% \\
& PrOntoQA & 20.63 [18.46, 23.05] & 0.228 & 91.8\% \\
& GSM8K & 96.66 [69.39, 122.27] & 0.548 & 71.2\% \\
\midrule
\multirow{3}{*}{\textbf{Gemma-2-9B}} 
& Dyck-$n$ & 12.44 [9.76, 14.60] & 0.422 & 82.0\% \\
& PrOntoQA & -52.48 [-57.98, -46.54] & 0.254 & 91.7\% \\
& GSM8K & 61.46 [51.51, 73.64] & 0.568 & 74.2\% \\
\bottomrule
\end{tabular}
\end{table}

This section reports extended NLDD, RSA, TAS, accuracy, and probability-delta results for all evaluated models. Figures are grouped by model for readability.

\begin{figure*}[b]
    \centering
    \begin{subfigure}[b]{0.32\textwidth}
        \centering
        \includegraphics[width=\linewidth]{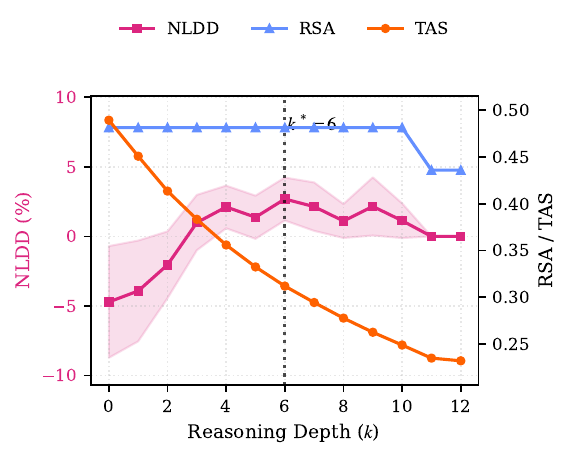}
        \caption{Dyck-$n$}
        \label{fig:appendix_llama_dyck}
    \end{subfigure}
    \hfill
    \begin{subfigure}[b]{0.32\textwidth}
        \centering
        \includegraphics[width=\linewidth]{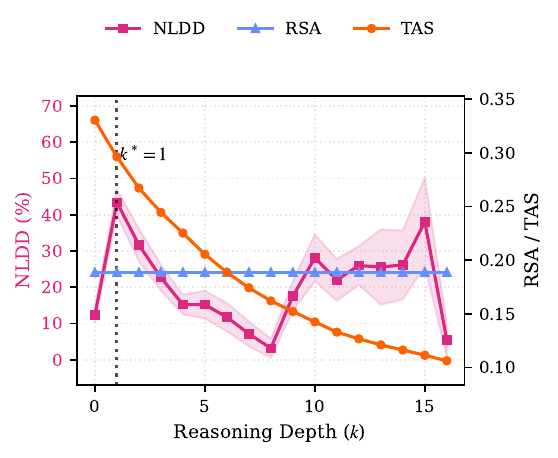}
        \caption{PrOntoQA}
        \label{fig:appendix_llama_prontoqa}
    \end{subfigure}
    \hfill
    \begin{subfigure}[b]{0.32\textwidth}
        \centering
        \includegraphics[width=\linewidth]{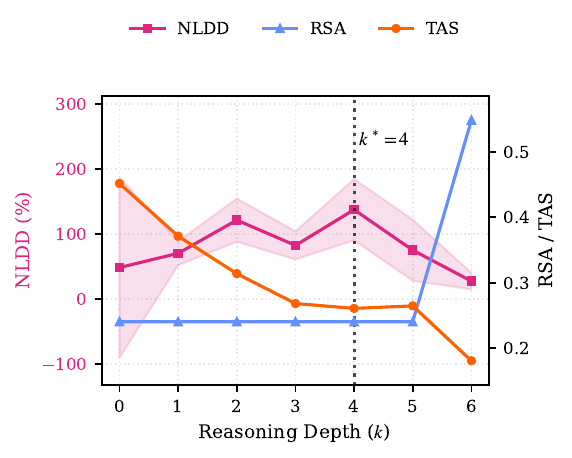}
        \caption{GSM8K}
        \label{fig:appendix_llama_gsm8k}
    \end{subfigure}
    \caption{LLaMA-3.1-8B: NLDD, RSA, and TAS as a function of corruption step index $k$ across tasks.}
    \label{fig:appendix_llama_nldd_rsa_tas_all}
\end{figure*}

\begin{figure*}[!t]
    \centering
    \begin{subfigure}[b]{0.48\textwidth}
        \centering
        \includegraphics[width=\linewidth]{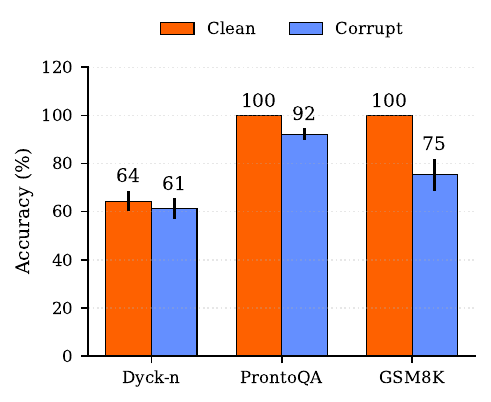}
        \caption{Accuracy under corruption}
        \label{fig:appendix_llama_accuracy}
    \end{subfigure}
    \hfill
    \begin{subfigure}[b]{0.48\textwidth}
        \centering
        \includegraphics[width=\linewidth]{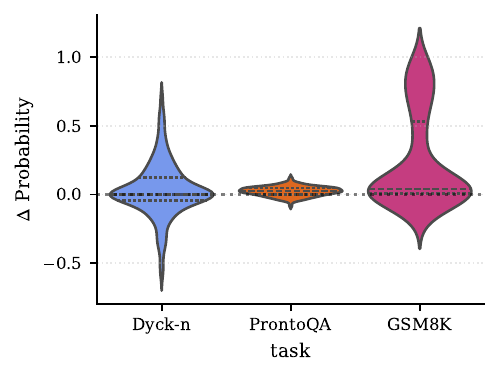}
        \caption{Probability delta distribution}
        \label{fig:appendix_llama_prob_dist}
    \end{subfigure}
    \caption{LLaMA-3.1-8B: robustness diagnostics under counterfactual step corruption.}
    \label{fig:appendix_llama_robustness}
\end{figure*}

\FloatBarrier
\vspace{0.5em}


\begin{figure*}[!t]
    \centering
    \begin{subfigure}[b]{0.32\textwidth}
        \centering
        \includegraphics[width=\linewidth]{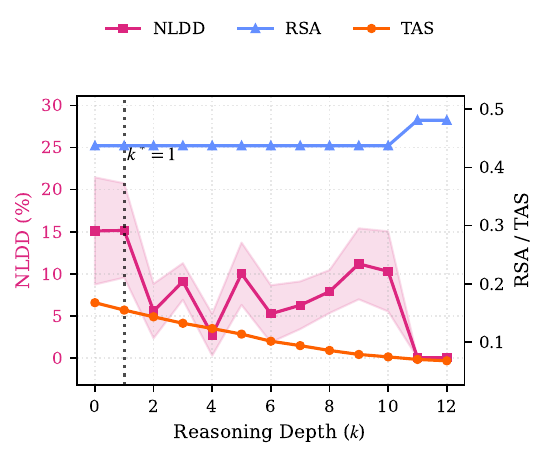}
        \caption{Dyck-$n$}
        \label{fig:appendix_deepseek_dyck}
    \end{subfigure}
    \hfill
    \begin{subfigure}[b]{0.32\textwidth}
        \centering
        \includegraphics[width=\linewidth]{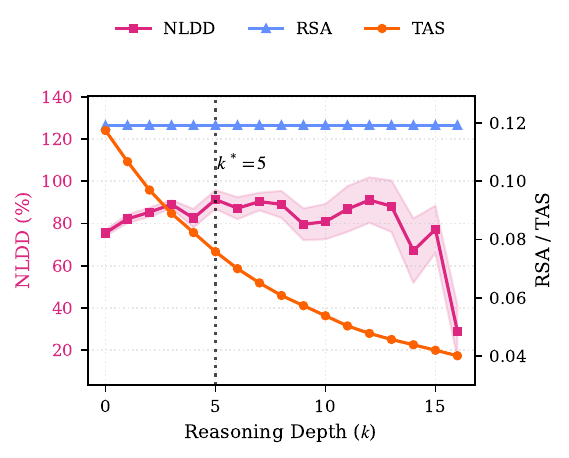}
        \caption{PrOntoQA}
        \label{fig:appendix_deepseek_prontoqa}
    \end{subfigure}
    \hfill
    \begin{subfigure}[b]{0.32\textwidth}
        \centering
        \includegraphics[width=\linewidth]{figures/fig1_horizon_gsm8k.pdf}
        \caption{GSM8K}
        \label{fig:appendix_deepseek_gsm8k}
    \end{subfigure}
    \caption{DeepSeek-Coder-6.7B: NLDD, RSA, and TAS as a function of corruption step index $k$ across tasks.}
    \label{fig:appendix_deepseek_nldd_rsa_tas_all}
\end{figure*}

\begin{figure*}[!t]
    \centering
    \begin{subfigure}[b]{0.48\textwidth}
        \centering
        \includegraphics[width=\linewidth]{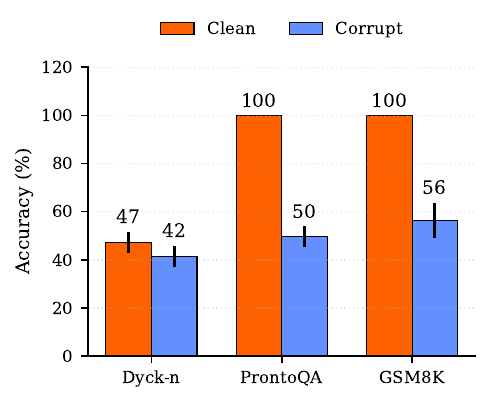}
        \caption{Accuracy under corruption}
        \label{fig:appendix_deepseek_accuracy}
    \end{subfigure}
    \hfill
    \begin{subfigure}[b]{0.48\textwidth}
        \centering
        \includegraphics[width=\linewidth]{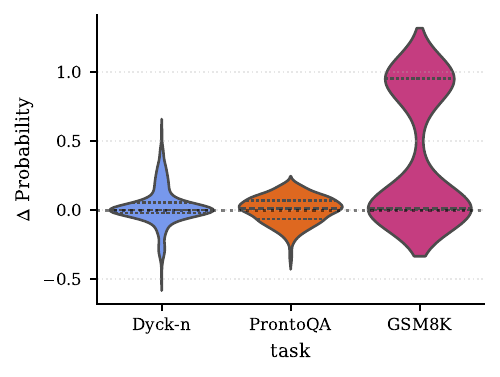}
        \caption{Probability delta distribution}
        \label{fig:appendix_deepseek_prob_dist}
    \end{subfigure}
    \caption{DeepSeek-Coder-6.7B: robustness diagnostics under counterfactual step corruption.}
    \label{fig:appendix_deepseek_robustness}
\end{figure*}

\FloatBarrier
\vspace{0.5em}


\begin{figure*}[!t]
    \centering
    \begin{subfigure}[b]{0.32\textwidth}
        \centering
        \includegraphics[width=\linewidth]{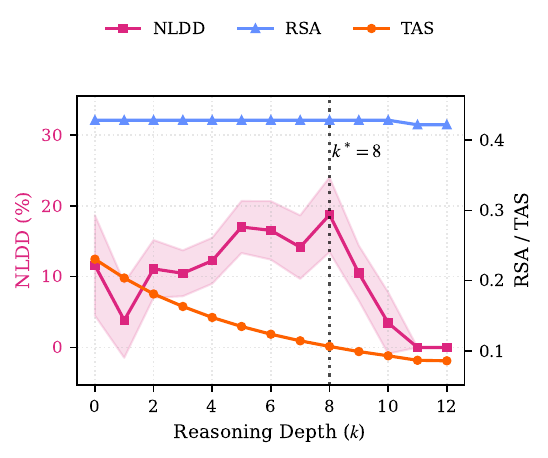}
        \caption{Dyck-$n$}
        \label{fig:appendix_gemma_dyck}
    \end{subfigure}
    \hfill
    \begin{subfigure}[b]{0.32\textwidth}
        \centering
        \includegraphics[width=\linewidth]{figures/fig1_horizon_prontoqa_g.pdf}
        \caption{PrOntoQA}
        \label{fig:appendix_gemma_prontoqa}
    \end{subfigure}
    \hfill
    \begin{subfigure}[b]{0.32\textwidth}
        \centering
        \includegraphics[width=\linewidth]{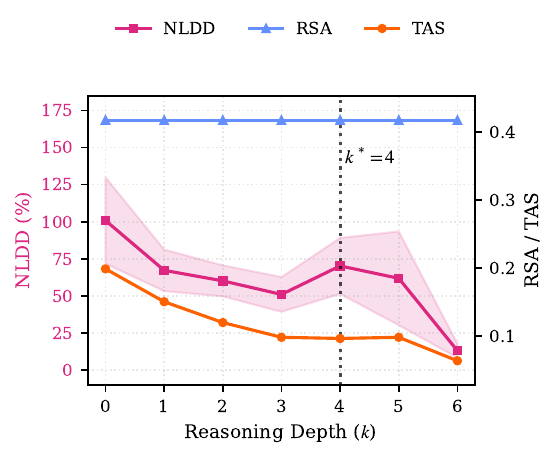}
        \caption{GSM8K}
        \label{fig:appendix_gemma_gsm8k}
    \end{subfigure}
    \caption{Gemma-2-9B: NLDD, RSA, and TAS as a function of corruption step index $k$ across tasks.}
    \label{fig:appendix_gemma_nldd_rsa_tas_all}
\end{figure*}

\begin{figure*}[!t]
    \centering
    \begin{subfigure}[b]{0.48\textwidth}
        \centering
        \includegraphics[width=\linewidth]{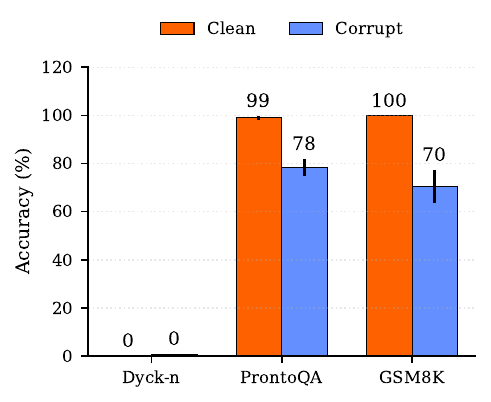}
        \caption{Accuracy under corruption}
        \label{fig:appendix_gemma_accuracy}
    \end{subfigure}
    \hfill
    \begin{subfigure}[b]{0.48\textwidth}
        \centering
        \includegraphics[width=\linewidth]{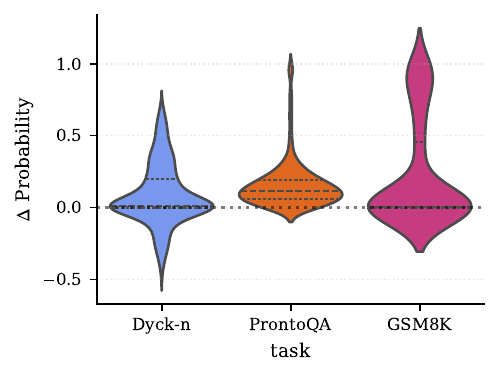}
        \caption{Probability delta distribution}
        \label{fig:appendix_gemma_prob_dist}
    \end{subfigure}
    \caption{Gemma-2-9B: robustness diagnostics under counterfactual step corruption.}
    \label{fig:appendix_gemma_robustness}
\end{figure*}

\FloatBarrier

\end{document}